\documentclass[letterpaper, 10 pt, conference]{ieeeconf}

\IEEEoverridecommandlockouts                              
                                                          
\usepackage{times}
\usepackage{helvet}
\usepackage{courier}

\usepackage{latexsym}
\usepackage[english]{babel}
\usepackage{graphicx}
\usepackage{verbatim}

\usepackage{amsmath}
\usepackage{amsfonts}
\usepackage{relsize}

\usepackage{amsmath} 
\usepackage{amssymb}  
\usepackage{amsfonts}

\usepackage{algorithm}
\usepackage{algpseudocode}
\usepackage{epstopdf}
\usepackage{listings}

\usepackage{longtable}
\usepackage{textcomp}
\usepackage{url}
\usepackage{alltt}
\usepackage{caption}
\usepackage{subcaption}
\usepackage{graphicx}
\usepackage{cite}

\usepackage[table,dvipsnames,svgnames]{xcolor}

\DeclareMathOperator*{\argmin}{arg\,min}

\newcommand{\myeq}[2]{\begin{equation}\label{eq:#1}\begin{aligned}#2\end{aligned}\end{equation}}

\newsavebox{\allttbox}

\DeclareMathOperator*{\argmax}{arg\,max}

\begin{document}


\title{Do What I Want, Not What I Did: \\ Imitation of Skills by Planning Sequences of Actions*}
\author{Chris Paxton$^{1}$, Felix Jonathan$^{1}$, Marin Kobilarov$^{1}$, and Gregory D. Hager$^{1}$
\thanks{*This work was supported by NSF NRI 1227277.}
\thanks{Chris Paxton, Felix Jonathan, Marin Kobilarov, and Gregory D. Hager are with the Laboratory for Computational Sensing and Robotics, Johns Hopkins University, Baltimore, MD 21218 USA. {\tt \small \{cpaxton3,fjonath1,marin,hager\}@jhu.edu}}}

\maketitle
\begin{abstract}
  We propose a learning-from-demonstration approach for grounding actions from expert data and an algorithm for using these actions to perform a task in new environments. Our approach is based on an application of sampling-based motion planning to search through the tree of discrete, high-level actions constructed from a symbolic representation of a task. Recursive sampling-based planning is used to explore the space of possible continuous-space instantiations of these actions.
  We demonstrate the utility of our approach with a magnetic structure assembly task, showing that the robot can intelligently select a sequence of actions in different parts of the workspace and in the presence of obstacles.
  This approach can better adapt to new environments by selecting the correct high-level actions for the particular environment while taking human preferences into account.
\end{abstract}

\section{Introduction}

\begin{figure}[bt!]
  \centering
  \includegraphics[width=0.99\columnwidth]{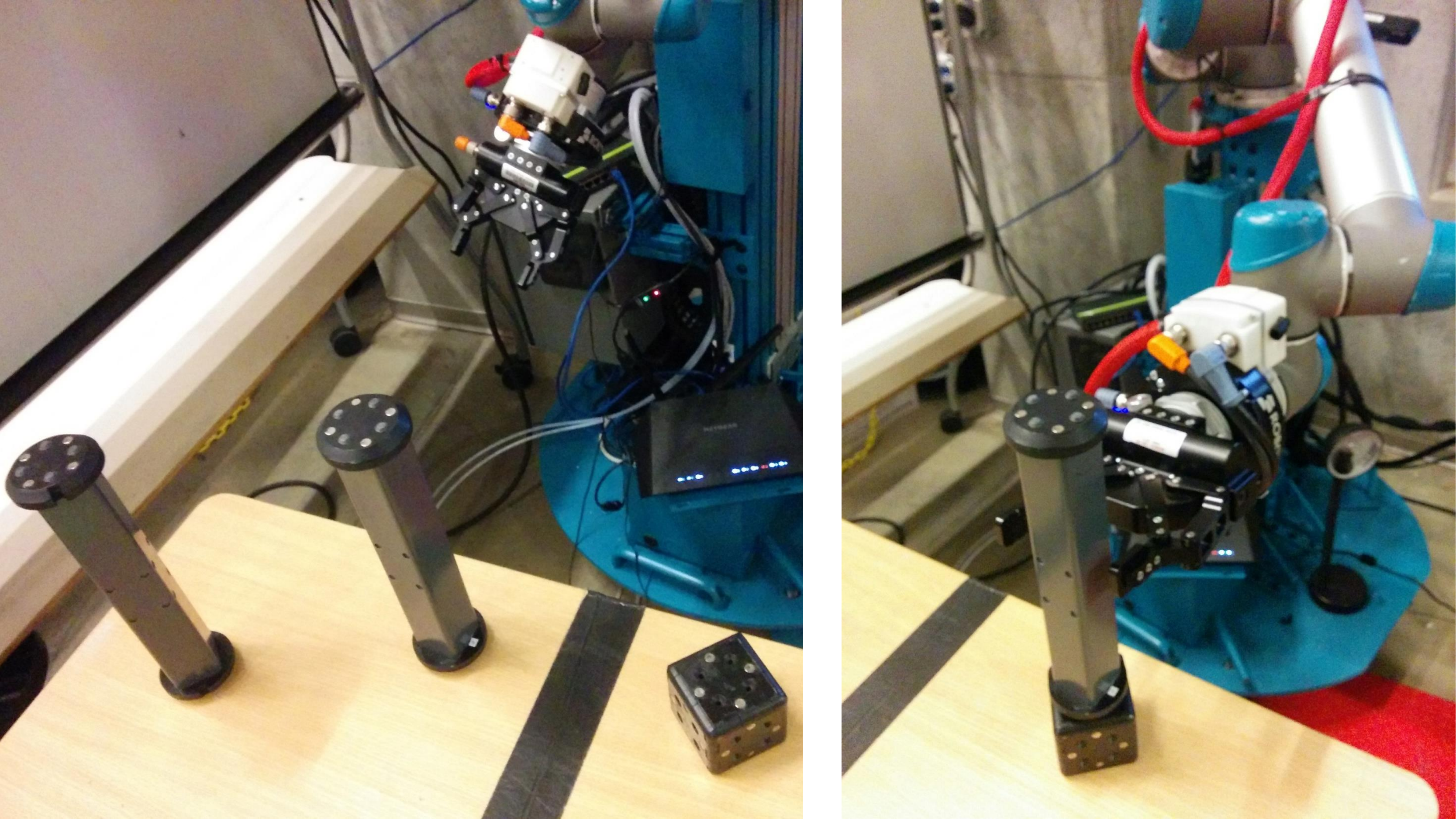}
  \caption{UR5 performing part of a structure assembly task by grabbing a link object in order to connect it to a node.
  Actions and goals were defined by human demonstrations.}
  \label{fig:cover}
\end{figure}

Learning from demonstration has emerged as a useful paradigm to teach robots the skills they need to interact with the real world. The challenge in learning from demonstration is to generalize what is learned to new contexts and new tasks.
Consider a moderately complex task such as assembling part of a structure, shown in Fig.~\ref{fig:cover} and defined by the PDDL in Fig.~\ref{fig:structure-pddl}.
The precise movements and the particular movement goals and parameters will vary from one situation to the next.
When attempting to execute this task in a new environment, the robot must be able to select the particular actions, motions, and manipulation goals that will allow completion of the task in this new environment.
By exploiting learned models for actions, we are able to demonstrate a planner that is able to produce solutions for performing tasks in an effective manner, is able to improve with additional demonstration data, and can adapt to new circumstances.

Adapting to new environments in the context of task and motion planning poses several challenges when attempting to generalize learned actions.
Recently there has been significant progress in integrating symbolic task planning and continuous motion planning~\cite{plaku2010sampling,wolfe2010combined,shivashankar2014towards,lagriffoul2014efficiently,plaku2015motion,bajada2015temporal},
which have in the past evolved as two separate fields.
At the same time, learning from demonstration has been established as a powerful tool for learning models of individual actions~\cite{argall2009survey,pastor2013dynamic}.
Learning from demonstration has previously been connected to symbolic task planning~\cite{kruger2009formal,wachter2013action,ahmadzadeh2015learning}, but these approaches are yet to be incorporated in the context of a motion-constrained task planning in a principled manner.
This paper aims to address this gap.

\begin{figure}[bt!]
\centering
\includegraphics[width=0.99\columnwidth]{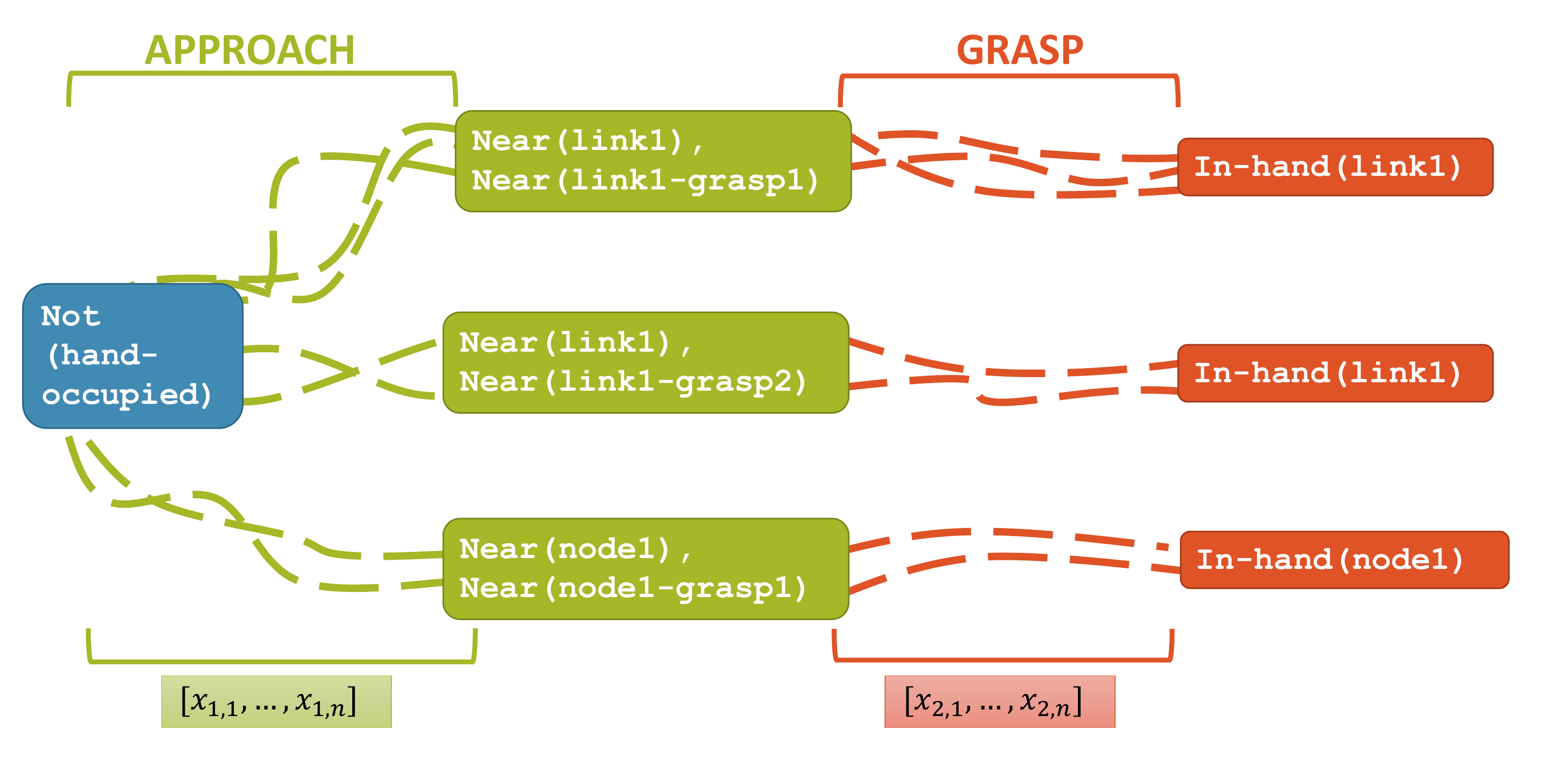}
\caption{Approach for grounding actions in symbolic planning. Training data represents actions connecting symbolic states in the graph of possible actions that constitute valid solutions to the task plan.}
\label{fig:data}
\end{figure}

In our approach, probabilistic models over features associated with each action are learned from human demonstrations and later refined in a supervised manner using additional robot-generated examples scored by a human teacher.
At the core of this approach lies a mapping from symbolic actions (e.g. \texttt{approach}, \texttt{grasp}) to physical motions encoded probabilistically as a distribution over observed features along each motion trajectory. Fig.~\ref{fig:data} shows this relationship: multiple demonstrations connect predicate states, which allow us to learn a model of each action.
Features $x$ are defined as  a set of functional relations between the robot and its environment (e.g. relative position and orientation between robot end-effector and desired object to grasp). 

Planning in a new environment is accomplished by updating each action distribution to remain as close as possible to the prior while satisfying new environment constraints such as different obstacles and object shapes.
This is accomplished through importance sampling and optimal distribution re-estimation using the cross-entropy method~\cite{rubinstein2004cross,kobilarov2011cem}.
Transitions from symbolic states to actions are similarly encoded as a discrete probability distribution representing the ``preference'' of executing different actions.
A product model is induced over a complete task from the sequence of probabilistic action models, together with discrete transition models.
Planning a complete task then corresponds to optimally updating this model to reproduce the prior and satisfy the new scenario. 

\begin{figure}[bt!]
\begin{lstlisting}[basicstyle=\scriptsize]
(define (domain structure-assembly)
  (:requirements :typing :adl)
  (:types link node grasp-pt)
  (:predicates
    (colliding) (hand-occupied)
    (near ?x - link) (near-grasp ?g - grasp-pt)
    (in-hand ?x - link) (standing ?x - link)
    (aligned ?x - link ?y - node)
    (grasp-feasible ?g - grasp-pt)
    (grasp-for ?x - link ?g - grasp-pt)
    (grasp-for ?y - node ?g - grasp-pt)
    (attached ?x - link ?y - node)
    (approach-feasible ?x - link))
  (:action approach
    :parameters (?x - link)
    :precondition (and
      (not (in-hand ?x))
      (approach-feasible ?x))
    :effect(near ?x))
  (:action grasp ...)
  (:action align ...)
  (:action place ...)
  (:action release ...)
  (:action disengage ...))

(define (problem build-simple-structure)
  (:domain structure-assembly)
  (:objects
    link1 - link link2 - link
    node1 - node node2 - node
  (:goal (exists (?x - link ?y - node) (attached ?x ?y))))
\end{lstlisting}
\caption{Partial PDDL domain and problem definition for the structure assembly task.
The domain can be thought of as a version of the basic blocks world task, where the goal is to latch 
two pieces together.}
\label{fig:structure-pddl}
\end{figure}

The contributions of this paper are: (1) a new method for reproducing demonstrated actions in novel environments, derived from sampling-based motion planning;
(2) an algorithm for combining these learned actions for executing multi-step tasks with multiple valid plans;
and (3) experimental validation of this algorithm on a simple assembly task as shown in Fig.~\ref{fig:cover}.
Experiments in a 2D Android game domain were omitted for reasons of space.

\section{Related Work}

Prior work exists in describing the relationship between high- and low-level actions and in learning representations of actions from demonstration, but does not combine learning with action selection and motion planning.

Object-Action Complexes (OACs) have been proposed as a way of formalizing actions unifying perception and learning that can be associated with learned low-level actions, and sequenced based on predicate effects by a symbolic planner~\cite{wachter2013action}.
The proposed method is similar to work such as~\cite{welke2013grounded}, which grounded PDDL position predicates with Gaussian Mixture Models, and~\cite{ahmadzadeh2015learning}, which associated Dynamic Movement Primitives (DMPs) for particular actions with expected visual features.

Probabilistic models are commonly used in imitation learning, e.g.~\cite{calinon2007learning,dong2011motion}.
Dynamic Movement Primitives (DMPs) are a policy representation that has proven useful for modeling low-level actions from demonstration as a set of dynamical systems~\cite{schaal2006dynamic}.
Prior work has added object avoidance to these methods through potential fields~\cite{park2008movement,ghalamzan2015incremental} or through reinforcement learning~\cite{kormushev2010robot}.

Pastor et al. used Path Integral Policy Improvement with DMPs and multiple human demonstrations to learn a model of expected features when executing two robotic tasks in~\cite{pastor2011skill}: shooting pool and flipping over a box with a pair of chopsticks.
This method was expanded upon by Stulp et al., who proposed Path Integral Policy Improvement with Covariance Matrix Adaptation~\cite{stulp2012path}.
These techniques are closely related to the Cross-Entropy Method for motion planning~\cite{kobilarov2011cem} from which we draw inspiration.

Our work is also related to the method proposed by Engbert et al. use the KL divergence between an expert demonstration and trajectories sampled from a Gaussian Process forward model to optimize imitation learning policies~\cite{englert2013model}.
Similarly in~\cite{boularias2011relative} the authors propose a method for inverse reinforcement learning based on minimization of relative entropy.
In addition, the proposed approach can be thought of as a parameterized set of actions; this has been shown to improve performance on policy learning in Markov Decision Processes~\cite{masson2015reinforcement}.

Other work combines learned motion primitives into state machines for execution, but in a purely reactive way: selecting only the next action, rather than the next sequence.
Examples include Niekum et al.~\cite{niekum2013incremental}, who build a task plan from unstructured demonstrations.
Manschitz et al.~\cite{manschitz2014learning} learned classifiers to determine the next action when sequencing motion primitives.
Work by Kappler et al.~\cite{kappler2015data} uses Associative Skill Memories to perform dexterous tasks.

Symbolic task planning and motion planning have commonly been integrated through algorithms that ``fill in the gaps'' in symbolic plans with callouts to continuous-space motion planners.
Recent work in combined task and motion planning include work by by Plaku et al.~\cite{plaku2010sampling}, by Shivashankar et al~\cite{shivashankar2014towards}, by Wolfe et al~\cite{wolfe2010combined}, and by Lagriffoul et al.~\cite{lagriffoul2014efficiently}. These works do not analyze actions with a wide variety of goals and cost functions, focusing instead on exploration and pick-and-place tasks. Similarly, Srivastava et al.~\cite{srivastava2014combined} efficiently integrate task planning with continuous-space reasoning about goal positions, but still rely on callouts to a traditional motion planner to instantiate trajectories. Work by Toussaint describes a hierarchal approach for integrated task and motion planning that first examines feasible end states before optimizing kinematics and motion planning~\cite{toussaint2015logic}. Unlike these methods, our approach jointly optimizes sequences of trajectories by adaptively allocating trajectory simulations to different actions. However, the proposed approach suggests directions for future work in improving efficiency in complex domains.

\section{Task Description}\label{sec:task-description}


We assume existence of (1) a symbolic description of a task, and (2) labeled training data associating features with each low-level action that can appear in this domain.
The symbolic description naturally decomposes the task into a sequence of predicate world states $w_0, w_1, ..., w_g$
For the structure assembly task, part of the symbolic description is shown in Fig.~\ref{fig:structure-pddl}. A world state $w$ is then defined as a combination of predicates. In turn, actions $a$ are the connections between these predicate states as shown in Fig.~\ref{fig:data}.
Each $a$ in a given task is represented as a probability distribution over a set of features associated with a successful instantiation of a skill in a new environment given $w$.

The features are denoted by $x\in\mathbb{R}^n$ and defined using the function $\phi$ through relationship
$x = \phi(t,s,u),$
where $t\in[t_0,t_f]$ denotes time in the action starting at $t_0$ and ending at $t_f$,
$s \in S$ is the robot state, and $u \in U$ are the applied controls.
With these definitions, a probabilistic model associated to each action $a$ is denoted by $p_d(x|a)$ and is computed using unsupervised learning from expert demonstrations, typically assuming a parametric density $p_d$. A joint model of a task $T$ consisting of multiple actions can be constructed using a density $p_d(x|T) \propto p_d(x|a_0)\cdots p_d(x|a_{n_T})$ assuming conditional independence between actions.

Specific features are derived from the PDDL description of the task.
For example, in Fig.~\ref{fig:structure-pddl}, the \texttt{approach} action describes the arm moving to pick up a \texttt{link} object without knocking it over.
In this case $x = \phi(t,s,u)$ would return the relative position, orientation, and velocity between the robot end effector and the \texttt{link} object.
To use the proposed method, one would provide the identifier for an action and a list of associated symbols from perception.

An optimal task $T^*$ is a sequence of actions $T^* = \{a_i\}_{i=1}^N$ that takes the robot from the initial state $w_0$ to goal $w_g$ that have the highest probability given our expert model, while also avoiding hard constraints such as collisions and joint limits:
\begin{align}
  T^* &= \argmax_T p(T | w_0, w_g) \\
  &= \argmax_{a_1,\dots,a_N} \prod_i^N p (a_i | w_i, w_g ) \label{eq:task}
\end{align}

Our goal is to learn a stochastic ``symbolic'' policy $\pi(a | w)$ over the sequence of predicate states, as well as continuous-space ``physical'' policy $p(u|s,\xi_a)$ generating trajectories for each action $a$.
We represent trajectories using parameters $\xi\in\mathcal Z$, where $\mathcal Z$ represents the space of all possible parameters resulting in valid trajectories in the new environment. Since robot perception and motion are uncertain, each parameter induces a density $p(\tau|\xi)$ where
\[
  \tau=\{\left<t_0,s_0,u_0\right>,\left<t_1,s_1,u_1\right>,\dots,\left<t_N,s_N,u_N\right>\}
\]
denotes the system trajectory. For instance, $\xi$ would typically define a reference trajectory and an associated tracking control law resulting in the density
\[
p(\tau|\xi) = p(s_0)\prod_{i=0}^{N-1} p(s_{i+1} | s_i, u_i) p(u_i | s_i, \xi).
\]
In practice, given $\xi$ the trajectory $\tau$ will either be sampled using a high-fidelity simulator or generated by the real robot.

\section{Planning Algorithm}\label{sec:algorithm}

\subsection{Local Planning Algorithm}\label{sec:local}

First, we consider adaptation of only a single action $a$ to a new environment.
When presented with a new environment, we pose the planning task as the problem of learning a new parameterized policy $\xi^*$.
To do so we employ a stochastic optimization technique using a surrogate distribution $\xi \sim \pi(\cdot| v)$ which is iteratively updated so that generated trajectories $\tau$ produce feature observations $x$ with high likelihoods under the expert distribution $p_d(x | a)$ for action $a \in A(w)$, where $A(w)$ is the set of actions available from predicate state $w$.


We follow the Cross Entropy Method described by Rubinstein et al.~\cite{rubinstein2004cross}, particularly following its application to motion planning by Kobilarov~\cite{kobilarov2011cem}. This is accomplished by introducing an artificial \emph{surrogate} distribution over $\mathcal V$ that will induce a distribution
over trajectories $\tau$ and over the corresponding features $x$ along these trajectories.
The surrogate will then be iteratively optimized until it becomes optimally close (in a distribution sense) to the expert density $p_d(x | a)$
without violating the constraints of the environment such as obstacles and joint limits.
The surrogate model is built using a parametric density $\pi(\xi|v)$ such as a multivariate Gaussian or a GMM with parameters $v$. 
Assuming that a nominal (prior) parameter $v_0$ is known the problem can be formalized as the optimal estimation of the expectation
\begin{align}
l &= E_{p(x | v_0)}[p_d(x | a)] \label{eq:l}.
\end{align}
The optimal importance sampling density~\cite{rubinstein2004cross} for estimating this integral is
\begin{align}
q^* &= \dfrac{p_d(x | a) p(x | v_0)}{l} \label{eq:optimal-q}
\end{align}
where the numerator in~\eqref{eq:optimal-q} can be thought of as the correlation between the expert feature distribution $p_d(\cdot | a)$ and the parameterized distribution $p(\cdot | v_0)$.
Unfortunately we cannot compute the solution to~\eqref{eq:optimal-q} as it involves computing the estimator $l$ from~\eqref{eq:l}.
Instead, we approximate this optimal $q^*$ by finding the appropriate parameters $v$ of $p(x | v)$.
A logical way of doing this is to minimize the Kullback-Leibler (KL) divergence:

\begin{align}
&\min_{v} D_{KL} (q^* || p (\cdot | v))
\end{align}

To find the value of $v$ that minimizes this expression, we approximate this solution by drawing $M$ i.i.d. samples $\xi_1,\ldots,\xi_M$ from $v_0$.
In this case $x_{i,j}=\phi(t_{i},s_{i,j},u_{i,j})$ is a generated feature from robot state $s_{i,j}$ at time $t_i$ along the sampled trajectory $\tau_j\sim p(\cdot | \xi_j)$ for $\xi_j\sim \pi(\cdot | v_0)$.

This can be more formally expressed as
\begin{align}
v^* = &\argmax_v \int_{x} p_d(x) p(x | v_0) \log p(x | v) \label{eq:kl2}\\
\approx &\argmax_v \dfrac{1}{N M} \sum_i \sum_{j} p_d(x_{i,j}) \log p(x_{i,j} | v) \label{eq:kl3}
\end{align}

If we assume that there is a bijection between a tuple $\left<t,s,u\right>$ along a trajectory $\tau$ and a feature $x\in\phi(\tau)$ then we have the following approximation
\begin{align}\label{eq:pxv5}
  p(x_{i,j}|v) \approx \dfrac{p(\xi_j | v)}{p(\xi_j | v_0)},
\end{align}
since $\xi_j$ were sampled under $v_0$,
and substituting~\eqref{eq:pxv5} into~\eqref{eq:kl3} results in:
\begin{align}
  &\argmax_v \dfrac{1}{NM} \sum_i \sum_{j} p_d(x_{i,j}) \log {p(\xi_j | v)}\label{eq:minkl}.
\end{align}

The necessary conditions for a minimum correspond to setting the gradient of~\eqref{eq:minkl} to zero, i.e. by solving the equality:
\begin{align}
  \sum_{i=0}^N \sum_{j=1}^M - z_{i,j} \nabla_v \log \pi(\xi_j | v) = 0 \label{eq:optcond},
\end{align}
where the weights $z_{i,j}$ are given by $z_{i,j} \triangleq p_d(x_{i,j})$.

When $\pi(\cdot |v) = \mathcal N(\cdot | \mu, \Sigma)\vert_{\mathcal V}$ (i.e. a single multivariate Gaussian with domain restricted to feasible parameter set $\mathcal Z$), the relationship~\eqref{eq:optcond} can be solved in closed form as
\begin{align}
\mu = \sum_{j=1}^M \bar z_{j} \xi_j, \quad \Sigma = \sum_{j=1}^M \bar z_{j} (\mu-\xi_j)(\mu-\xi_j)^T,
\end{align}
where $z_{j} = \sum_{i=0}^N z_{i,j}$ and $\bar z_j = z_j/\sum_{j=1}^M z_j$.
When $\pi(\cdot |v)$ is a GMM the minimization from Eq.~\eqref{eq:minkl} is performed using a weighted expectation maximization (EM) algorithm. 

In practice, the optimal parameter $v$ is computed iteratively starting with some nominal choice $v_0$ which approximately covers the trajectory space of interest.
At each iteration we draw $M$ samples $\xi_j \sim \pi(\cdot|v_0), j \in 1,\dots,M$  and compute the next $v$ by minimizing~\eqref{eq:minkl}.
At the next iteration $v_0$ is set to $v$ and the process continues until the cost converges. 


We add a fixed normalization term to the diagonal entries in $\Sigma$ of $p_d$ and of $\pi(\cdot | v)$ to make sure covariances stay well-defined.
In addition, to prevent premature convergence, we introduce an extra parameter $0 < \alpha < 1$, which controls the size of steps taken at each iteration.
In the case where $v$ is multivariate Gaussian, with $\Sigma_i^*$ as the optimal $\Sigma$ at iteration $i$,
we compute $\mu_{i+1}$ and $\Sigma_{i+1}$ as:
\myeq{step-size}{
	\mu_{i+1} &= (1 - \alpha)\mu_i - \alpha \mu^*_i \\
	\Sigma_{i+1} &= (1 - \alpha)\Sigma_i - \alpha \Sigma^*_i
}



\subsubsection*{Avoiding Obstacles and Joint Limits}
We constrain $\mathcal{Z}$ to consist only of the space of valid trajectories, removing any samples that would collide with objects or pass joint limits.
This means that when drawing our $M$ samples, we remove samples currently in collision or past joint limits in our new environment and continue to draw sample trajectories until we have all $M$ valid examples.
This works effectively in practice as long as the task does not require generalization in environments with very narrow passages that the system has never been trained on. Such cases are extremely difficult since the probability of obtaining samples in the narrow passage is close to zero, unless an informative nominal density parameter $v_0$ is used with enough probability mass over such regions. 

\subsection{Task Planning Algorithm}
We wish to optimize parameters for all possible actions in a successful execution of the task,
where our cost is the joint probability over any sequence of actions that represent a valid execution of the task as per Eq.~\eqref{eq:task}.
Our task planning approach takes the algorithm described in Section~\ref{sec:local} and expands it into a recursive algorithm similar to Monte Carlo Tree Search.

%

First, consider the problem of choosing one of $N_{A(w)}$ possible actions.
We think of this as the choice of which action would be most similar to our expert's demonstrations in other scenes, starting in symbolic state $w$. 
We expand our notion of $p_d(x)$ to include the switch between each possible action as $p_d(x) = p_d(a_i | w) p_d(x_{i,j} | a_i)$. 
Substituting this into Eq.~\eqref{eq:minkl} gives us:
\begin{align}
    &\argmax_v \dfrac{1}{NM} \sum_j^M p_d(a_j | w)\sum_{i}^N p_d(x_{i,j} | a_j) \log {p(\xi_j | v_{a_j})}\label{eq:minkl}.
\end{align}
where $v_{a_i}$ is the trajectory distribution associated with $a_i$.

Action selection is modeled as a stochastic policy over possible worlds.
We introduce a surrogate distribution into our trajectory search that captures the probability of choosing each future action from the current $w$. When sampling trajectories, we draw the next action $a \sim \pi(\cdot | w)$ according to this probability.

Furthermore, we can extend this reasoning to consider which of a whole tree of possible actions is the most similar to an expert tree, allowing us to capture expert preferences for particular actions in addition to continuous-space trajectories.
Assuming that all actions in a branch of the tree are independent given time, 
we can define the expert probability of a particular action starting at continuous robot state $s_0$: 

\begin{align}
 Q(w,s,a) &= \dfrac{1}{NM} \sum_i^N \sum_j^M p_d(x_{i,j} | a) V(W(a), s_{N,j}) \label{eq:Q} \\
 V(w, s) &= \sum_{a' \in A(w)} p_d(a' | w) Q(w, s, a') \label{eq:V}
\end{align}

Where $s_{N,j}$ is the final state in sampled trajectory $\tau_j$ and $w$ represents the world after symbolic action $a$. Eq.~\eqref{eq:V} describes the probability of all possible actions from a continuous world state $s$ occurring after execution of an action $a$.

When recording a set of $N_w$ demonstrations starting in the same predicate state $w$, we compute the conditional probability for action $a \in A(w)$:

\begin{align}
p_d(a | w) &= \frac{\sum_i^{N_w} I_{\{a = A_i\} }}{N_w}
\end{align}

We specify a surrogate distribution over possible choices of actions for a world $w$ given as $p(a|w)$.
This probability is initialized as
$\pi(a|w) = \frac{1}{N_{A(w)}}$. In the case where $H=0$ this is updated as $\pi(a|w) \propto \frac{1}{M}\sum_j^M z_j$
 where $M$ trajectory samples $\tau$ have been drawn from $a$.
 Otherwise we compute this as:
 \[ \pi(a|w) = \frac{p_d(a | w)}{M} \sum_{s_0} Q(s_0, a) \]
 for starting state $s_0 \in S_0$ and action $a \in A(w)$.
 In practice we use the step size $\alpha$ to prevent this term from converging too quickly.

Each predicate state $w$ corresponds to a range of valid continuous-space states.
The algorithm recursively samples from the trajectories associated with each successive action to map to continuous states.
As shown in Alg.~\ref{alg:complete}, we repeatedly call the {\sc Sample} function from Alg.~\ref{alg:forward}, providing it with the set of possible start states $S_0$. We select a start state from these $s_0$ according to the cumulative probability of these actions.
The process continues until we reach a user-provided horizon $H$.
This approach allows us to maintain a constant number of samples: over successive iterations, more samples will be devoted to promising regions of the search space.

\begin{figure}[bt!]
\centering
\begin{subfigure}[b]{\columnwidth}
\centering
\includegraphics[width=0.75\columnwidth]{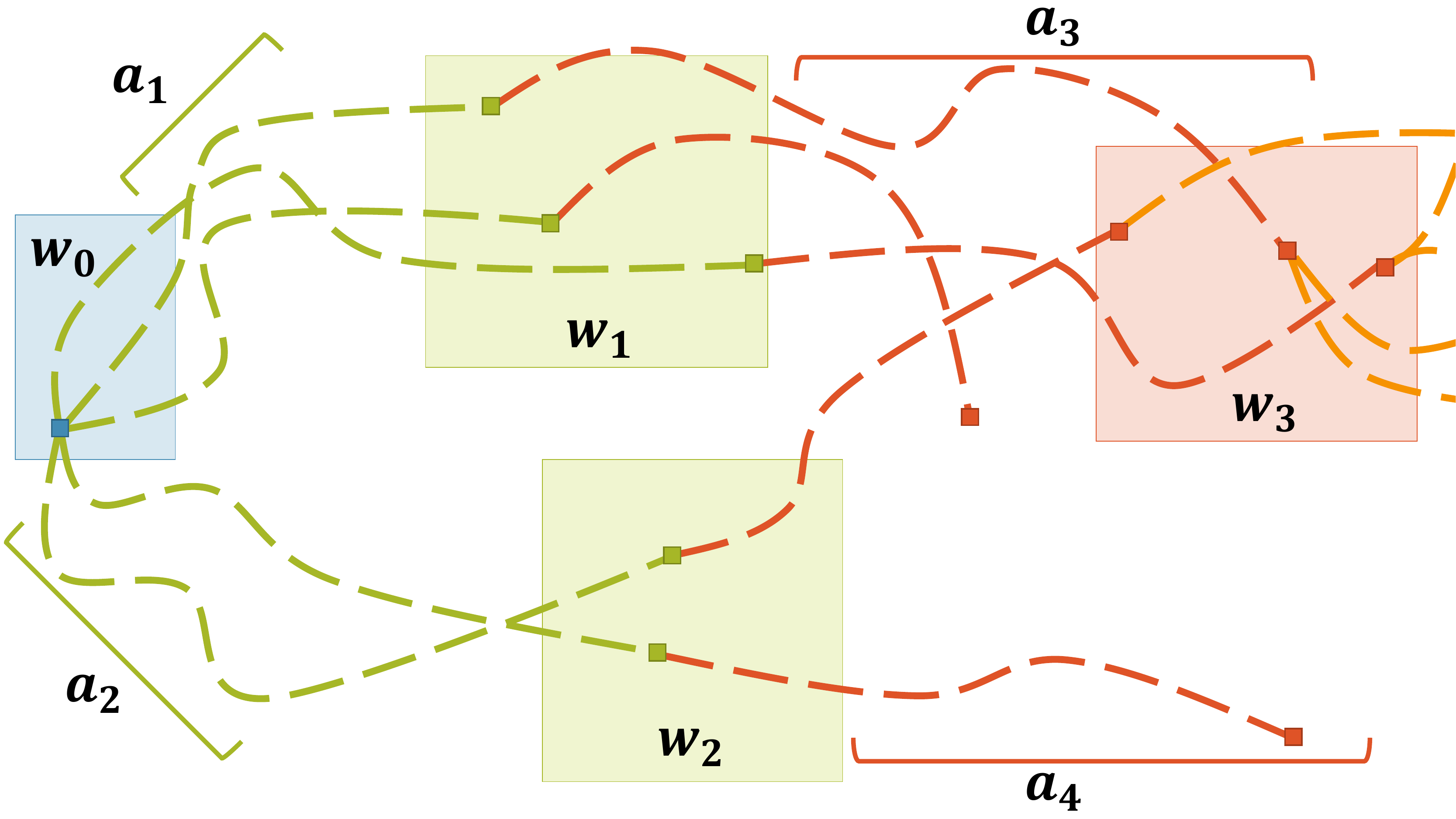}
\caption{At the first iteration of the algorithm, we sample trajectories (dashed lines) corresponding to $a_1$, $a_2$, $a_3$, etc. according to  and compute $p_d(\tau | a, w)$. Trajectory distributions for $\pi(\cdot|v_1)$,$\pi(\cdot|v_2)$, etc. are updated, as are $\pi(a | w)$}
\label{fig:alg-start}
\end{subfigure}
\begin{subfigure}[b]{\columnwidth}
\centering
\includegraphics[width=0.75\columnwidth]{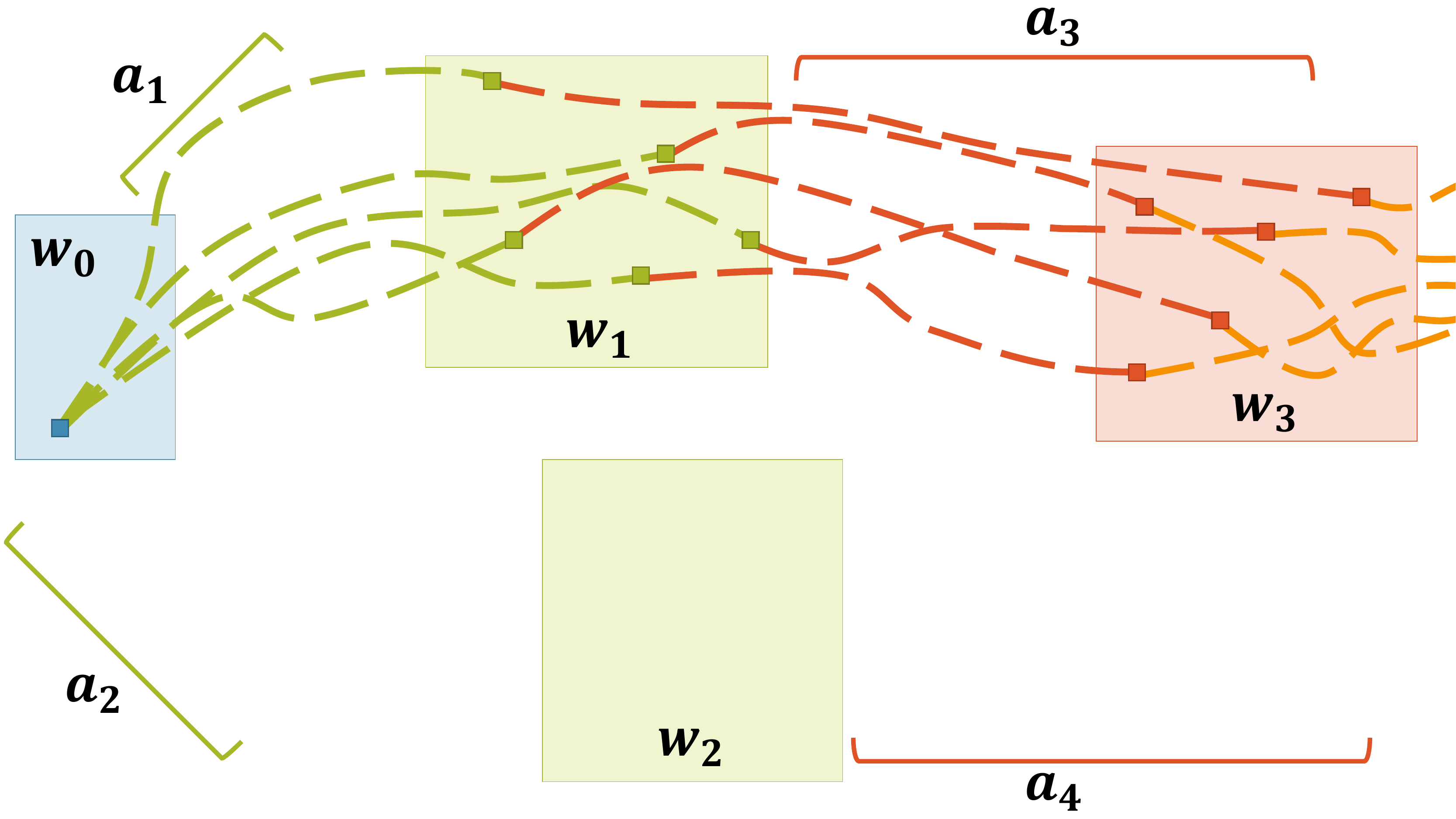}
\caption{On subsequent iterations, trajectory sampling is biased towards $a_1$ due to the comparatively high probability of valid trajectories for each action in this space.}
\end{subfigure}
\caption{Illustration of the proposed algorithm. Boxes $w_1$, $w_2$, etc. indicate regions corresponding to the predicate state after each action, while dashed lines represent continuous state trajectories.}
\label{fig:alg-steps}
\end{figure}

This results in a recursive search strategy outlined in Alg.~\ref{alg:complete}. The return of the {\sc Sample} function is the average probability of all future actions and trajectories associated with each current start state. This value is used to compute a version of the weights in Eq.~\eqref{eq:minkl}, where $p_d$ is replaced by the probability of all future actions from each trajectory.
Fig.~\ref{fig:alg-steps} illustrates how the algorithm works in practice.

\begin{algorithm}[bt!]
\caption{Pseudocode algorithm for optimal reproduction of demonstrated tasks in new environments.}
\label{alg:complete}
\begin{algorithmic}
\State Given: initial world $w_0$, initial state $s_0$, horizon $H$, step size $\alpha$, max iterations $N_{iter}$
\For  {$i \in N_{iter}$}
	\For {$a \in A(w_0)$}
		\State $Q(w_0, s_0, a) =$ \Call{Sample}{$a$,$s_0$,$1$,$H$,$M$}
	\EndFor
	\State $V(w_0, s_0) = \sum p_d(a | w_0) Q(w_0, s_0,a)$
	\If{$V(w_0, s_0)$ has converged} 
		\State Break;
	\EndIf
\EndFor
\end{algorithmic}
\end{algorithm}

\begin{algorithm}[bt!]
\caption{Recursive trajectory sampling and update step.}
\label{alg:forward}
\begin{algorithmic}
\Function{Sample}{$a,S_0,p(S_0),H,M$}
\State $w = W(a)$ \Comment Predicate world after performing action
\For {$j \in [1,...,M]$}
	\State $s_{0} \sim S_0 \propto p(S)$ \Comment Sample start points
	\State $\xi_{j} \sim \pi(\cdot | v_{a})$ \Comment Sample parameters
	\State $\tau_{j} \sim p(\cdot | \xi_j, s_{0})$ \Comment Compute trajectories
\EndFor
\If {$H > 0$}
	\State $S_0' = [s_{N}]_{j=1}^N$ 
		\Comment Set start points
	\State $p(S_0') = [p(s_{0,j})p_d(\tau_j | a)]_{j=1}^M$
	\State \Comment Compute probabilities of each start point
	\For {$a' \in A(w)$}
		\State $H' = H - 1$ \Comment update horizon
		\State $M' = \pi(a' | w) M$
			\Comment Number of samples
		\State $Q(S_0', a') =$ \Call{Sample}{$a'$
$S_0'$,$p(S_0')$,$H'$,$M'$}
		\State $\pi(a'|w) = \frac{p_d(a' | w)}{M'} \sum_{s_0'} Q(s_0', a')$
	\EndFor
\EndIf
\For {$j \in [1,\dots,M]$}
	\State $V(s_{N,j}) = \sum_{a' \in A(w)} p_d(a' | w) Q(s, a')$
	\State $z_{j} = \sum_i p_d(x_{i,j} | a) V(s_{N,j})$
	\State \Comment Compute update weights from child probability 
\EndFor
\For {$s_0 \in S_0$}
	\State $V(s_0) = \frac {\sum_j I_{s_{0,j} = s_0} z_j}{\sum_j I_{s_{0,j}}}$
	\State \Comment Average probability from continuous start state
\EndFor
\State $v_{a}' = \argmin_v \frac{1}{N} \sum_i \sum_{j} z_j \log {p(\xi_j | v_a})$
\State \Return $V(S_0)$
\EndFunction
\end{algorithmic}
\end{algorithm}

\section{Experiments}
We performed experiments in a simulated Barrett WAM arm and on a Universal Robot UR5, applied to an object manipulation task.
The goal of this task was to build a structure of increasing complexity out of magnetic blocks, as per the task described in~\cite{bohren2013pilot}. In our case, we only perform a part of the whole structure assembly task: we combine one link and one node object to create a sub-structure.
The connections between different skills are described by the PDDL specification in Figure~\ref{fig:structure-pddl}.
We used FastDownward~\cite{helmert2006fast} to translate the PDDL into a graph of possible actions that can be performed assuming all \emph{feasibility} predicates are true.

\begin{figure}
  \centering
  \includegraphics[width=0.99\columnwidth]{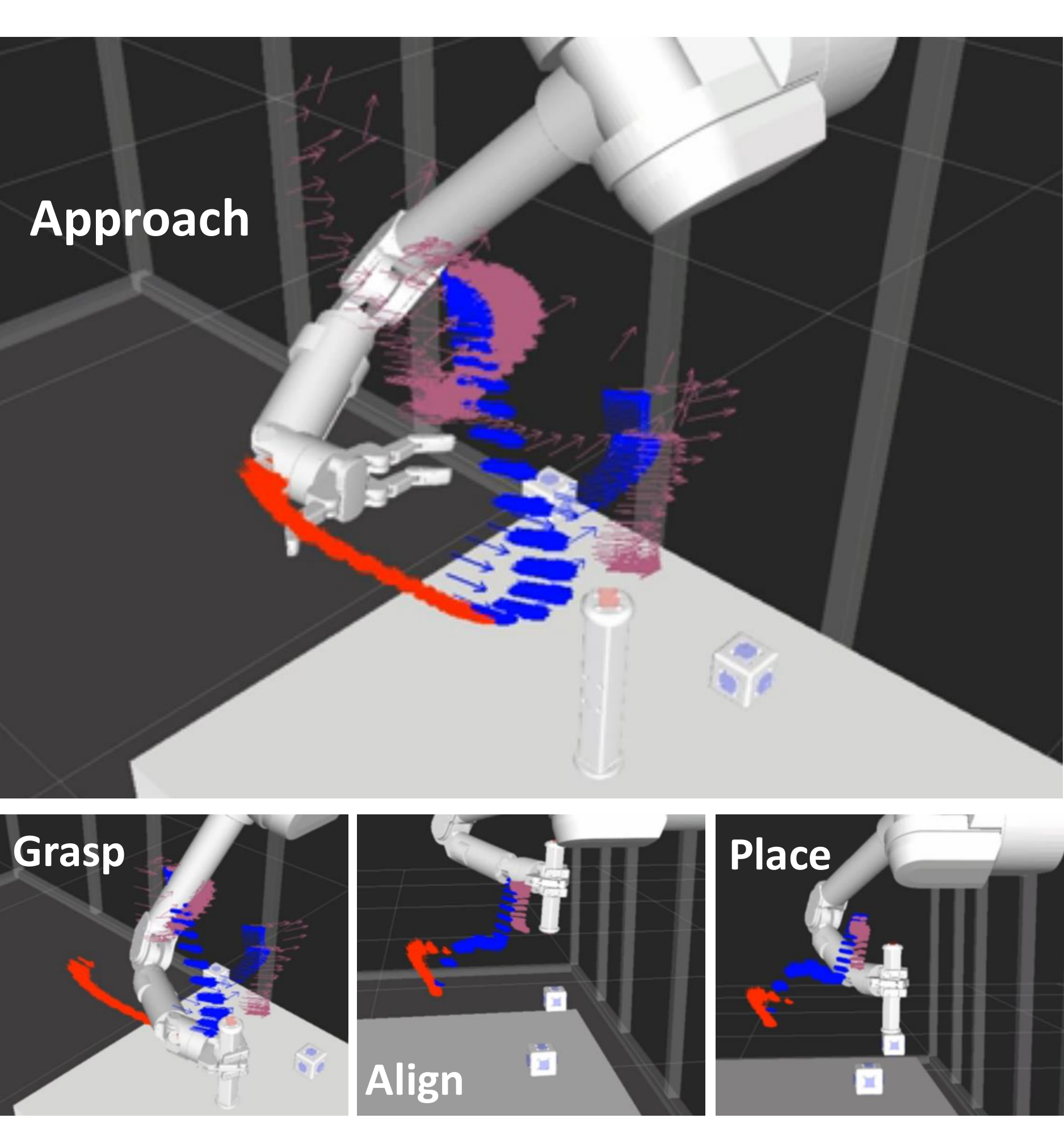}
  \caption{Graphic showing example plan for the simple structure assembly task discussed here. Different colors indicate \texttt{approach}, \texttt{align}, and \texttt{place} actions. The planner has selected the leftmost node, and chose to grasp the link from the right.}
  \label{fig:planning}
\end{figure}

Figure~\ref{fig:planning} shows how the planner works in practice.
It iteratively sampled out different motions for each selected action, choosing to approach the link from the front and then to mate it to the leftmost node.
As the algorithm progressed, successively fewer samples were drawn from actions associated with the rightmost node.

We use three types of features: (1) the time in a particular state, (2) the gripper command variables, (3) the transforms between the end frame and the objects. Relevant features are determined by the parameters specified in the task description in Fig.~\ref{fig:structure-pddl}.
In cases where two objects are parameters, we used the transform between the in-hand object and the other object.
For these examples, $\phi(t,s,u) = [t,p_x,p_y,p_z,r_x,r_y,r_z,r_w,\|p\|,\dot{p}_x,\dot{p}_y,\dot{p}_z,\|\dot{p}\|]$, where values are computed from the offset between the current manipulation frame and and the relevant object.
The values $(r_x,r_y,r_z,r_w)$ define a unit quaternion.
The manipulation frame is either an end effector position or the coordinate frame associated with the object in the gripper, for actions defined where the \texttt{hand-occupied} predicate is true.

We parameterize trajectories $\xi$ with Dynamic Movement Primitives with $5$ basis functions in the robot's joint space, plus a goal pose $g \in SO(3)$.
This allows us to find paths in the space of the robot arm, but to adapt to different possible continuous-space goals. 
We implemented the system using ROS~\cite{quigley2009ros} with Orocos KDL for inverse kinematics~\cite{orocos-kdl}.

In practice, the ``link'' object can shift in unpredictable ways after a grasp action, so we adjust the plan after completion of the grasp action.
We add noise to the parameters of the trajectory distribution associated with the subsequent \texttt{align} and \texttt{place} actions and replan.
In the real robot experiment we omit this step due to the lack of accurate position information once the object is in the gripper.

The current implementation of the planner is single-threaded.
The single largest inefficiency was detecting collisions, followed by computation of inverse kinematics.
Particularly in scenes with more obstacles, both of these are very important: inverse kinematics are required to adapt trajectories to different possible grasp points, and accurate collision detection guarantees safe execution.
As inverse kinematics and collision detection are outside the purview of this paper, we did not focus on efficiency.

\subsection{Simulation Experiments}\label{ssec:comparison}
We collected three demonstrations of each of the different skills with a dynamic simulation of the Barrett WAM arm.
We then place these pieces in different positions in the environment, and validated our method by performing the task in different locations.
The results of one performance in a novel environment are shown in Fig.~\ref{fig:planning}.

To create a model of each of these skills, we collect three demonstrations of the object manipulation action using the same grasp, with two of the Barrett Hand's fingers on the left side of the link and one on the right. The simulated WAM arm was teleoperated with a Razer Hydra to collect training data.
The user provided examples of three different grasps: a direct approach and approaching from the left or the right.
The user specified $p_d(\tt{approach}|w_0)$ to indicate a preference for a direct approach.

\begin{figure}[bt!]
  \centering
  \includegraphics[width=0.99\columnwidth]{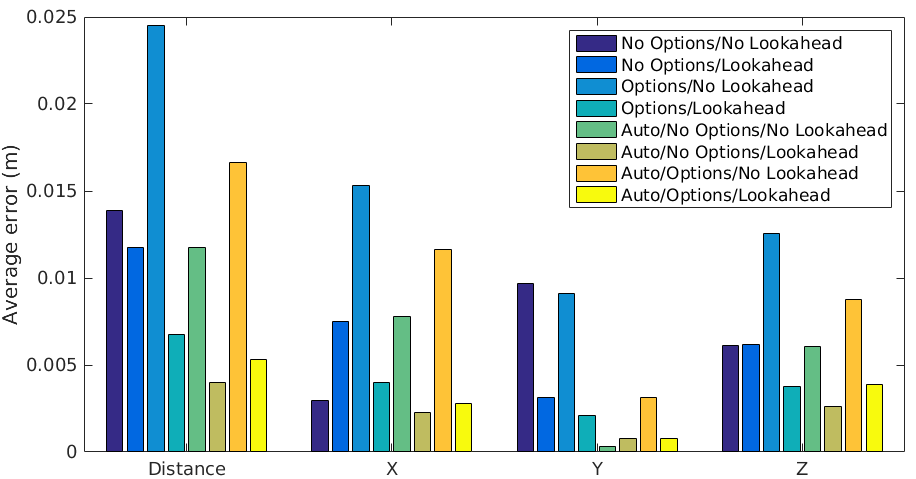}
  \caption{Plot showing absolute error in distance, x, y, and z from ``perfect'' mate position between link and the selected node. Our full algorithm ("Options/Lookahead") achieved high mate accuracy, roughly equivalent to the version with no options, and was able to complete the task in more challenging scenarios.
  }
  \label{fig:error}
\end{figure}

We perform our task on scenes with one link and two nodes at different positions, and demonstrate task effectiveness for $10$ trials with different configurations of the world.
The key measure of performance is how easily we can add extra training data to our model and how close these results will be to the target mate.
We set $\alpha=0.5$ and used $M=200$ trajectories, with a maximum of $15$ iterations. Our full algorithm used a depth of $H=5$: a long enough horizon to plan the whole assembly task.
Average likelihood of sampled trajectories converged exponentially as we proceeded through various iterations.
Fig~\ref{fig:error} shows distance to an ideal final mate after outliers were removed. 


By way of comparison, we remove one or both of two parts of our algorithm.
We use a single randomly selected task plan (``no options'' in Fig.~\ref{fig:error}).
We also compare against the case where our planner only examines the currently available actions, setting $H=1$ (``no lookahead'' in Fig.~\ref{fig:error}).
The ``no options/no lookahead'' case functions as our baseline: it uses the algorithm in Sec.~\ref{sec:local} to reproduce an action based on a GMM.
While our approach is technically unconstrained, due to the sampling method we implicitly constrain the trajectory search to a feasible set of valid trajectories.
To demonstrate how we can improve performance by improving action models, a human user selected three successful trials from the automatic performance of this task and added them to the model (``auto'' in Fig.~\ref{fig:error}).

Table~\ref{table:failures} shows the number of planning failures associated with different environments.
These are cases where the algorithm failed to find a trajectory with nonzero probability under the expert distributions defining each of our actions.
Without the full algorithm, either the robot often cannot find a solution that will accomplish the task or performance is significantly degraded.
The case where there are options and no lookahead is a good example. While the robot is almost always able to find a plan in this situation, the quality of plans is far worse, as shown by Fig.~\ref{fig:error}. In the higher-performing ``Auto'' case, the robot was always able to find a plan but few of these plans were successful: only 4/10 achieved high-quality mates, and several outliers fell off the node and the table completely. This is because without knowledge of the \texttt{place} and \texttt{release} actions, the \texttt{align} action will often not terminate in a good state to complete the task.

Fig.~\ref{fig:error} shows a comparison on successful trials in different environments without obstacles.
The full algorithm was highly reliable and accurate, achieving less than 1 cm of placement error.
Other versions of the algorithm made mistakes that planning alone could not recover from.
In addition, performance of all versions of the algorithm showed improvement when extra data was added, though the full version of the algorithm was still better and more flexible.

\begin{table}[bt!]
\centering
\begin{tabular}{| c | c c |}
\hline
& Expert Data Only & With Auto Data\\
\hline
No Options/No Lookahead & {\color{red} 7} & {\color{red} 4} \\
No Options/Lookahead & {\color{red} 3} & {\color{red} 1} \\
Options/No Lookahead & {\color{red} 1} & {\color{green} \textbf{0}} \\
Options/Lookahead* & {\color{green} \textbf{0}} & {\color{green} \textbf{0}}  \\
\hline
\end{tabular}
\caption{Number of failures when generalizing to novel environments. 
(*) indicates the full algorithm. Columns represent whether model was taught using only expert demonstrations or whether extra data was added from successful executions.}
\label{table:failures}
\end{table}

\begin{figure}
  \centering
  \includegraphics[width=\columnwidth]{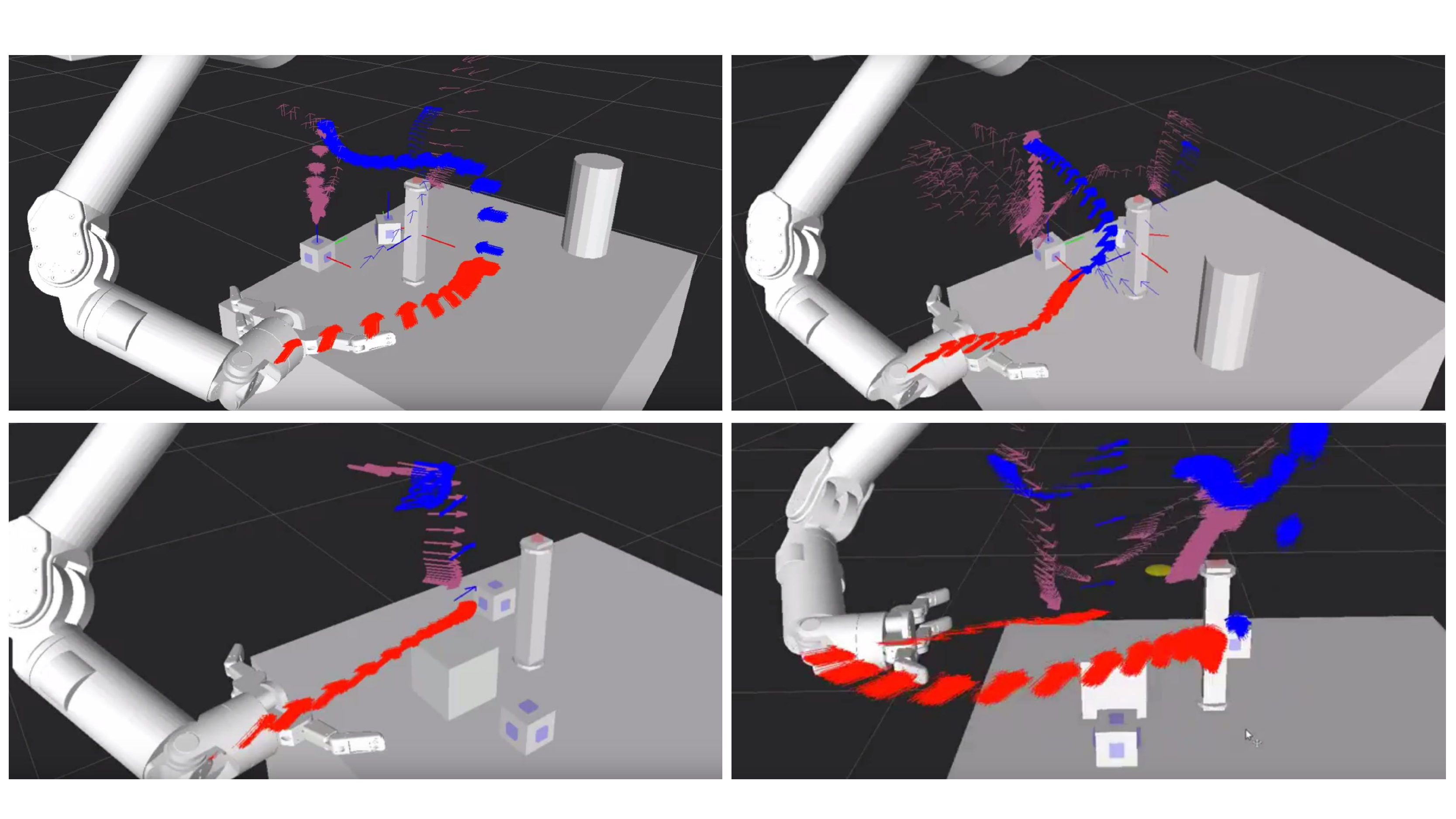}
  \caption{Performance of the planner in different environments with the addition of obstacles.
  The planner chooses paths that are consistent with taught actions as much as possible.}
  \label{fig:avoidance}
\end{figure}

We also introduced different obstacles into the environment. In these cases, the algorithm is able to avoid these objects and still complete its required task.
Figure~\ref{fig:avoidance} shows examples of these results.
Since the planner removes paths that are in collision, this restricts the set of feasible trajectories.
As in the upper left of Figure~\ref{fig:avoidance}, the most likely action in a particular scenario might be an approach from a particular direction. Once this grasp is blocked, the planner can either attempt a less likely trajectory that results in that grasp, or it can approach from a different direction.
This tradeoff illustrates why our approach is more powerful than adding a potential field term to action primitives as in~\cite{park2008movement} or similar work.

\subsection{Real Robot Experiments}

The UR5 was given the option of grabbing either of two links or a node object and combining them to create the same structure as in the simulation experiments.
Demonstrations were provided in which the robot grasped either the node or the link first.
The object localization technique described by Li et al.~\cite{li2016hierarchical} was used to determine the poses of all objects in the scene.

Our system intelligently selected whether to grasp the link or the node, and selected which face of the link to grasp based on feasibility and presence of other obstacles.
The UR5 had a fairly limited workspace and has a limited ability to interact with objects when compared to the Barrett WAM arm used in the simulation experiments, making this a more challenging problem.
However, the robot was able to grasp both node and link objects and complete the task. Our video supplement provides examples of the UR5 performing this task in different configurations, as well as an overview of the algorithm and videos of the simulation experiments.

In particular, when presented with both link and node objects in different orientations, the robot was able to correctly select available faces not blocked by other obstacles. If the node was better aligned with the robot's gripper, then the algorithm chose to grasp the node; if one of the links was better aligned, it would grasp this link.


\section{Conclusions}

We described a practical approach for task and motion planning based on models of skills grounded from expert demonstrations of skills.
By representing actions as probability distributions learned from expert demonstrations, we create a framework that can combine a broad range of actions to accomplish a task. We validated this approach with experiments in a structure assembly domain both in simulation and in a real robot.
While we did not address efficiency in the implementation used in this paper,
we will examine strategies for decreasing the number of costly trajectory evaluations and collision checks and apply our planner to larger and more complex tasks.

\bibliographystyle{../style/IEEEtran}
\bibliography{../bib/machine_learning,../bib/taskmodels,../bib/lfd,../bib/software,../bib/planners,../bib/surgery,../bib/task_description,../bib/us,../bib/vision}

\end{document}